# Explainable Semantic Text Relations: A Question-Answering Framework for Comparing Document Content

**Yehudit Aperstein [1], Alon Gottlib[1], Gal Benita[1] and Alexander Apartsin[2]**

[1] Intelligent Systems, Afeka Academic College of Engineering, 218 Bnei Efraim St., Tel-Aviv
[2] School of Computer Science, Faculty of Sciences, Holon Institute of Technology (HIT), 52 Golomb St., Holon 5810201, Israel

**Abstract**

Understanding semantic relations between two texts is crucial for many information and document management tasks, in which one must determine whether the content fully overlaps, is completely superseded by another document, or overlaps only partially, with unique information in each. Beyond establishing this relation, it is equally important to provide explainable outputs that specify which pieces of information are present, missing, or newly added between the text pair.

In this study, we formally define semantic relations between two texts through the set-theoretic relation between their respective Answerable Question Sets (AQS), the sets of questions each text can answer. Under this formulation, Semantic Text Relation (STR), such as equivalence, inclusion, and mutual overlap, becomes a well-defined set relation between the corresponding texts' AQSs. The set differences between the AQSs also serve as an explanation or diagnostic tool for identifying how the information in the texts diverges.

Using this definition, we construct a synthetic benchmark that captures fine-grained informational relations through controlled paraphrasing and deliberate information removal supported by AQS manipulations. We then use this dataset to evaluate several discriminative and generative models for classifying text pairs into STR categories, assessing how well different model architectures capture semantic relations beyond surface-level similarity. We publicly release both the dataset and the data generation code to support further research.

**Keywords:** Semantic text relation benchmark; Entailment; Semantic similarity; Large Language Models; Answerability; Question Answering, Synthetic Data; Textual Inference

## 1. Introduction

Understanding semantic content relations between texts is essential for many information and document management applications, where it is critical to determine whether textual information fully overlaps, is completely superseded in one document by another, or overlaps only partially with unique content in each. In information retrieval, identifying whether documents in the retrieved set offer redundant or complementary content is essential for assembling a complete, non-redundant answer to a user's query. In multi-document summarization, such relations enable systems to separate shared from unique information, producing more concise and accurate summaries. The ability to determine whether one text conveys the same information as another, or semantically subsumes it, is essential in many domain-specific settings. For example, legal and regulatory analysis depends on detecting when a new document paraphrases, extends, or diverges from existing policy language to ensure compliance and correct interpretation. In such contexts, explainable semantic text relations are crucial: practitioners must not only know that two documents differ but also understand exactly how they differ. Highlighting which information is added, missing, or altered enables precise identification of changes in obligations, protections, or compliance requirements, reducing the risk of misinterpretation or oversight.

In domains where information may be presented in inherently incomplete or noisy ways, this challenge becomes even more pronounced. Recent work shows that large language models can recover meaning and identify missing information in free-form clinical narratives through question–answer reasoning [1]. Such reasoning directly motivates the need for a formal framework capable of representing what information is present, absent, or overlapping across texts. Beyond healthcare, similar reasoning patterns have emerged in education, where generative AI systems simulate teacher–student interactions to identify conceptual gaps between a learner's response and the reference explanation [2].

Identifying Semantic Text Relations (STR) is difficult because natural language is flexible, context-dependent, and often ambiguous. The main challenge lies in separating real semantic differences from superficial variation, where two texts may convey the same information using different wording, structure, or distribution of facts across sentences.

Even after the relation is determined, explaining it poses an additional challenge: one must isolate the specific informational units that differentiate the texts. A complete content difference must show what Text A conveys that Text B does not, what Text B conveys that Text A does not, and what information they share despite differing phrasing.

We focus on three foundational semantic relationships: semantic equivalence, semantic inclusion, and semantic overlap. Semantic equivalence holds when two texts convey the same underlying information, even if expressed with different wording or spread across multiple sentences. Semantic inclusion applies when all the information in one text is fully contained in another, regardless of their relative lengths or structure. Semantic overlap captures cases in which two texts share some, but not all, of their informational content, with each contributing unique elements alongside a common core. Together, these relationships offer a structured view of how texts relate at the level of conveyed meaning, independent of phrasing or surface form.

Semantic Text Relations (STR) formalize and generalize well-known Natural Language Inference (NLI) relations such as entailment, contradiction, and neutrality by grounding them in Answerable Question Sets (AQS). In NLI, entailment means that the content of one text logically follows from, or is guaranteed by, the information in another, and this judgment is typically made subjectively by a human reader. STR expresses this more precisely: Text A entails Text B if every question answerable from B is also answerable from A, meaning that A's AQS fully contains B's AQS. Moreover, Natural Language Inference (NLI) typically operates on single, factual statements, where entailment focuses on explicit facts or direct inferences within short, self-contained inputs. STR extends this notion to complex texts, where information may be dispersed across multiple sentences, implicitly stated, or expressed at different levels of abstraction.

Recent QA-based evaluation frameworks, such as AlignScore [3], QAPyramid [4], and PlainQAFact [5], share the underlying assumption that semantic content can be operationalized through the set of questions a text can answer. However, these works primarily employ QA procedures as evaluation metrics for specific downstream tasks, such as summarization or factual consistency. In contrast, our framework formalizes this principle as a general semantic model: two texts are semantically related if their sets of answerable questions (AQS) are equivalent, nested, or intersecting. Thus, while inspired by the same QA-based semantic evaluation paradigm, the proposed STR framework extends it from task-specific evaluation to task-agnostic semantic representation and benchmark construction.

This AQS-based, set-theoretic perspective unifies both NLI-style entailment and QA-based evaluation under a single formal representation of meaning.

The set-based view enables extending beyond the standard NLI labels by introducing additional relation types, such as overlap, which indicates that two texts answer some of the same questions while each also contains information the other lacks. STR therefore offers a broader and more nuanced taxonomy of relations between texts. In this work, we focus on informational content, but the same framework can easily incorporate relations such as contradiction by examining questions for which the texts give semantically different answers.

The task of producing an *explainable* STR also becomes clearer under the AQS framework. Differences between the AQS of two texts directly reveal how their content diverges. A question answerable from Text A but not from Text B identifies information present only in A. A question answerable only from B shows what B adds or changes. In this way, STR provides both a formal relation and an explicit explanation of the content differences that give rise to it.

In this work, we create the first synthetic benchmark for estimating STR directly from texts. The dataset contains texts together with their corresponding AQS, as well as paired examples labeled according to the relations derived from their AQS. In principle, the AQS of a text can contain an infinite number of questions that differ only in wording. In practice, we approximate each AQS with a representative sample of questions drawn from real question-answering datasets such as SQuAD. We also use SQuAD as a starting point for generating the texts themselves, ensuring that both the questions and the associated passages reflect naturally occurring information content.

Using SQuAD passages as base material, we generate rich synthetic datasets of document pairs through controllable paraphrasing and systematic information removal, producing examples that exhibit semantic equivalence, inclusion, or overlap. Generative models further produce diverse, contextually appropriate questions, enabling robust test sets that cover a broad range of semantic content. To ensure a reliable gold standard, we pre-filter and post-filter examples to retain only those for which a state-of-the-art LLM can consistently answer questions about the information that should remain in the text after paraphrasing and removal.

We use a generative LLM (GPT 4.1) to manipulate source texts by prompting it to paraphrase passages or remove specific pieces of information in a controlled manner guided by the target AQS. We also rely on the LLM for both pre-filtering and post-filtering of the generated texts to ensure that the resulting passages correspond precisely to the specified AQSs.

As a first step, we use the resulting benchmark to evaluate several classification approaches that rely solely on text pairs. The explanatory component, in which a model would extract the relevant AQS questions that represent the differences or overlaps between the texts, is left for future work.

We compare the ability of a generative model (GPT-4.1 and GPT-4o) to classify text pairs into STR types in zero-shot and few-shot prompting settings. In addition, we fine-tune several pretrained transformers (DistilBERT, BERT, RoBERTa, and Longformer) and evaluate their performance as supervised STR classifiers. We also train traditional machine learning models, such as Random Forests and Logistic Regression, using statistical features extracted from each text pair.

This work makes the following contributions:

1. *A formulation of explainable Semantic Text Relations (STR) grounded in Answerable Question Sets (AQS)*: We define the semantic relation between two texts through the set-theoretic relationship between their respective sets of answerable questions. Comparing these sets determines whether the texts are semantically equivalent, one is included in the other, or they partially overlap. Because each difference in answerability corresponds to a specific piece of missing or added information, this approach provides a clear, operational, and explainable definition of STR based on LLM-driven answerability.
2. *A first STR benchmark dataset*: We construct and release a synthetic dataset of text pairs labeled with fine-grained semantic relations and corresponding question sets, which can be used to build and evaluate explainable STR models.
3. *Evaluation of discriminative and generative STR classifiers*: We train and compare multiple approaches for directly classifying semantic text relations from text pairs, including zero-shot and few-shot prompting with large generative models, supervised transformer-based classifiers, and traditional machine-learning models.

## 2. Prior Work

### *2.1 Textual Semantic Relations*

Semantic relations between sentence pairs, such as entailment (one sentence implying another), paraphrase (meaning equivalence), contradiction, and graded similarity, are fundamental to language understanding. Early work framed Recognizing Textual Entailment (RTE) as a unified evaluation task: given two text fragments, determine the other [6]. Modern practice often uses the terms Natural Language Inference (NLI) and RTE interchangeably, with tasks usually classifying a premise-hypothesis pair as entailment, contradiction, or neutral [7]. For instance, Bowman et al. [8] note that "understanding entailment and contradiction is fundamental to understanding natural language". Paraphrase detection treats the problem as a binary classification: deciding whether two sentences are semantically equivalent [9]. Semantic Similarity tasks assign a graded score (often 0–5) reflecting the degree to which two sentences are alike: for example, the SemEval STS tasks measure meaning similarity on a continuous scale [10]. Contradictions when sentences express mutually incompatible facts are typically modeled as the negation of entailment in NLI. In sum, the broad goal is to recognize whether one sentence entails or contradicts another, whether two are paraphrases, or how similar their meanings are. These tasks overlap; for example, paraphrases are bidirectional entailments and typically rank high in similarity, whereas contradictions receive low similarity scores.

Before the advent of deep learning, methods for these tasks were primarily symbolic or logic-based. Early RTE systems relied on lexical and syntactic features, including string overlap, WordNet relations, edit distance, and syntactic parse alignment [11, 12]. Some approaches performed sentence alignment or graph matching to compare premise and hypothesis tokens or dependency structures [12,13]. Logical and rule-based inference was also explored; for example, [14] developed a logic-theorem prover, NaturalLI, that mapped text to first-order logic forms, and [15] created natural logic rules to derive entailments without requiring complete logical forms. In [16] survey, they note that solutions "*range from shallow approaches based on lexical similarities... to advanced methods that consider syntax, perform explicit sentence alignment or use formal logic*". Lexical databases, such as WordNet, enabled path-based similarity [17] and synonym-based entailment checks. Distributional semantics (LSA, word embeddings) also played a role; for instance, [18] demonstrated how vector similarity can capture paraphrases. However, these classical models often struggled with compositional meaning or required heavy feature engineering.

A variety of annotated corpora have driven research. The PASCAL RTE challenges [6] provided small sets of sentence pairs labeled for entailment (yes/no). The SNLI corpus (Stanford NLI; [8]) revolutionized the field by providing ~570,000 human-authored English sentence pairs labeled as entailment, neutral, or contradiction. This scale enabled the training of data-hungry neural models. The MultiNLI corpus [19] extended SNLI to ~433,000 pairs drawn from ten different written and spoken genres, allowing evaluation of cross-genre generalization. The SICK dataset [20] comprises approximately 10,000 English sentence pairs, annotated for both semantic relatedness (on a 5-point similarity rating scale) and categorical entailment (entailment, contradiction, or neutral). For paraphrase identification, the Microsoft Research Paraphrase Corpus [9] contains ~5,800 news-sentence pairs labeled as paraphrases or not. More recently, PAWS (Paraphrase Adversaries from Word Scrambling) provides 108K high-lexical-overlap sentence pairs labeled paraphrase/non-paraphrase, specifically to challenge models that ignore word order and structure. Semantic similarity benchmarks include the SemEval STS shared tasks (2012–2017) and the STS Benchmark [10], which measure correlation between model scores and human ratings. The GLUE benchmark [21] combines multiple tasks, including MNLI, RTE, QQP, MRPC, and STS-B, among others, into a single evaluation suite. Finally, for multilingual NLI, XNLI [22] extends MultiNLI's development and test sets into 15 languages (e.g., Arabic, Spanish, Urdu), enabling cross-lingual inference evaluation. These datasets have standardized evaluation: typically, classification accuracy (NLI, paraphrase) and Pearson/Spearman correlation for regression similarity tasks.

With large datasets available, neural networks quickly came to dominate the field. Early neural NLI models encoded each sentence (typically using a BiLSTM or CNN) into a vector and then combined or compared the vectors for classification. Bowman et al. [8] reported a baseline in which separate LSTM encoders for the premise and hypothesis achieved ~77.6% accuracy on SNLI. Soon after, more sophisticated architectures appeared. Wang et al. [23] introduced match-LSTM, an LSTM that processes the hypothesis while attending to the premise at each word; this model achieved an accuracy of 86.1% on SNLI, substantially outperforming earlier baselines. Attention mechanisms became central: Parikh et al. [24] proposed a Decomposable Attention model that explicitly aligns ("soft-attends") words or phrases between the two sentences before comparing them. Their simpler, attention-based model achieved state-of-the-art performance on SNLI with significantly fewer parameters. Similarly, Chen et al. [25] developed the ESIM (Enhanced Sequential Inference Model), which uses BiLSTMs and a bidirectional attention step; ESIM achieved a score of 88.6% on SNLI, the highest to date. These models all follow a familiar pattern: attending to align substructures, comparing aligned pieces, and aggregating comparison features (the "Attend-Compare-Aggregate" framework). Parikh et al. [23] note that for NLI it often suffices to "simply align bits of local text substructure and then aggregate" rather than building a single global embedding.

For paraphrasing and similarity, similar Siamese or attention-based networks have been employed. Mueller and Thyagarajan [26] trained Siamese LSTM networks to map sentences into a shared embedding space and used cosine similarity for paraphrase detection. More recently, sentence-encoder approaches have thrived; for example, InferSent [22] utilized supervised NLI data (SNLI) to train a universal encoder, demonstrating that it outperformed unsupervised embeddings on a range of tasks. Reimers and Gurevych [27] proposed Sentence-BERT (SBERT), which involves feeding each sentence through a shared BERT encoder, a Siamese/triplet network, and computing cosine similarity between the fixed embeddings. SBERT dramatically speeds up similarity search (since each sentence is encoded once) while maintaining BERT-level accuracy on semantic textual similarity benchmarks. These methods contrast with cross-encoder models (like standard BERT fine-tuning on sentence pairs), which jointly encode both sentences and are slower at inference.

The most significant recent leap has come from large pre-trained Transformer models. Devlin et al. [28] introduced BERT ("Bidirectional Encoder Representations from Transformers"), which is pre-trained on vast text corpora with masked language modeling and next-sentence prediction objectives. Crucially, BERT can be fine-tuned on a downstream task by simply adding a classifier. Devlin et al. [28] demonstrated that fine-tuning BERT surpassed the state of the art on multiple tasks, with notable improvements. For instance, the MultiNLI accuracy increased to 86.7%, and the GLUE benchmark score rose from previous highs into the 80s. In effect, BERT learned rich contextual sentence representations that encode many facets of meaning and inference. Follow-up work revealed that larger and better-tuned variants outperformed BERT. Liu et al. [29] demonstrated that with more training data and hyperparameter tuning (RoBERTa), one can "match or exceed" all prior models and set new records on GLUE. Models such as XLNet [30] and ALBERT [31] employ alternative training objectives or parameter sharing to enhance performance further. As a result, modern fine-tuned transformers, such as BERT, RoBERTa, and XLNet, achieve near-human or superhuman results on standard NLI and similarity benchmarks. For example, BERT alone improved the GLUE score to 80.5 (+7.7 absolute points) and substantially reduced errors on entailment and similarity tasks. These models are now the dominant approach: virtually all state-of-the-art NLI or semantic similarity systems start with a pretrained transformer and fine-tune on the target dataset.

### 2.2 *Answerability*

The challenge of answerability arises when a QA system must determine whether a given question can be answered based on the context provided. Early QA benchmarks (like SQuAD v1.1) assumed all questions were answerable by extracting a span from the text. Modern QA datasets explicitly include unanswerable or "no-answer" questions to force models to learn when to abstain. For example, Rajpurkar et al. [32] introduced SQuAD 2.0 by adding ~50,000 crowd-sourced unanswerable questions to the original SQuAD data. These "impossible" questions are adversarially written to appear like answerable ones, so systems must not only find answers when they exist but also predict correctly when the passage supports no answer. Similarly, other benchmarks incorporate answerability judgments: Natural Questions (NQ) asks real user queries against Wikipedia and marks many as unanswerable; NewsQA includes "nil" questions (no answer in the article) due to how questions were generated; QuAC contains dialog questions that are often open-ended or unanswerable; and BoolQ consists of naturally-occurring yes/no questions (always answered "Yes" or "No" based on a paragraph).

English QA dominates this space. In addition to SQuAD 2.0 and NQ, NewsQA is a CNN-based news dataset for span extraction that includes a non-trivial fraction of unanswerable questions. QuAC [33] is a question-answer dialog dataset in which many questions lack answers or depend on dialog context. BoolQ [34] contains naturally occurring yes/no questions that require entailment-style reasoning. While all are answerable by design (each answer is either "yes" or "no"), BoolQ highlights that binary QA tasks also implicitly require answer verification. More recently, multilingual tasks like TyDi QA [35] have been extended to 11 languages. In its primary functions, the model must select either an answer span, a yes/no response, or report NULL if no answer is available. (In a simplified "gold passage" variant, TyDi discards unanswerable questions by construction.) Overall, these datasets spotlight answerability: a system must learn to output nothing (or a special token) when the evidence is insufficient.

QA models handle answerability mainly by integrating a no-answer prediction alongside answer extraction. Modern approaches can be grouped as follows:

Joint span-extraction with a "no-answer" score: Transformer-based extractive models (e.g., BERT) are typically extended to include an extra output for no-answer. For instance, BERT's QA head can produce a probability for "empty" answer (often by using the [CLS] token) and a special SoftMax that includes the no-answer option. In one design, the model computes both span-start and span-end scores, as well as a separate "no-answer" score, applying a SoftMax over all possibilities. Training then penalizes incorrect predictions of either spans or no-answer. In effect, the model jointly learns to extract an answer span when present and to flag unanswerable cases. Kundu and Ng [36] describe such a "nil-aware" model, which returns a span if one exists or outputs "Nil" otherwise. In experiments on NewsQA, this approach outperforms simpler pipelines and naive thresholding.

Pipeline vs. Threshold methods: An alternative is to decouple answer extraction and answerability detection. In a pipeline approach, one first applies an answer extractor (trained only on answerable data) and then uses a separate classifier to decide if the answer is valid. A threshold-based method, on the other hand, utilizes the model's confidence. After the span is predicted, if its probability (or some model confidence score) falls below the chosen threshold, the system returns a 'no-answer'. This threshold can be tuned on development data. Kundu and Ng report that simple thresholding (setting high-entropy predictions to 0) is inferior to end-to-end models, yet it remains a common baseline. Notably, Kamath et al. [37] emphasize that raw SoftMax confidences are often poorly calibrated: a high SoftMax score may not reflect true certainty, especially under domain shift. They propose training a separate "calibrator" to predict when the model is likely to be wrong, thereby abstaining more reliably than naive thresholding.

Answer verification ("read-then-verify"): Some methods explicitly double-check answers. For example, Hu et al. [38] developed a two-stage system: a reader extracts a candidate answer and outputs a probability that there is no answer, and a separate "answer verifier" verifies whether the context entails that answer. The verifier can be another neural model (often another transformer) that takes the question, context, and candidate answer and scores its plausibility. If the verifier finds the answer unsupported, the system abstains from processing. Such architectures have demonstrated strong performance: Hu et al. report a F1 score of ~74.2 on SQuAD 2.0 (state-of-the-art at the time) using their read-and-verify model; other work similarly ensembles a QA model with an entailment checker or auxiliary "answerability" head.

Generative models and output tokens: More recently, seq2seq transformers (e.g., T5, BART) have been fine-tuned for extractive question-answering (QA). These can also be trained to emit a special output when no answer is available. For instance, a T5-based QA model can be given a special classification token ([CLS]) at the start of the question; when fine-tuned on SQuAD2.0, it will output [CLS] (effectively "no answer") for unanswerable queries. In other words, the sequence model learns to generate either an answer span or a "no-answer" token. This approach turns answerability detection into a generation task. Likewise, more open-ended LLMs (e.g., GPT) can be prompted to respond with "I do not know" when unsure. However, ensuring reliability often requires careful fine-tuning or the development of effective prompting strategies, which are not yet fully resolved.

Transformer-era models have largely subsumed older QA architectures, but the need for answerability remains critical. Pretrained models, such as BERT [28] or T5, are typically extended with an additional head or token for the "no-answer" option. For example, BERT-based QA yields a span and also assigns a probability to "no answer", usually derived from the [CLS] token's output. In practice, teams fine-tune these models end-to-end on datasets like SQuAD 2.0. A well-tuned BERT can achieve ≈ 80–85% F1 on SQuAD 2.0, which is far below its performance on the all-answerable SQuAD 1.1, reflecting the difficulty of answerability detection. Models like UnifiedQA or MultiQA train on many QA formats (span, Boolean, multiple-choice) and inherently handle yes/no and no-answer cases across tasks. In all cases, the key is that the model must learn robust abstention behavior to avoid hallucinating answers for unanswerable queries.

Many key QA datasets incorporate answerability:

- SQuAD 2.0 (2018) dataset extends SQuAD 1.1 by adding 50K unanswerable questions. These negatives are adversarially created, so that an answerable-looking question may have no answer in the paragraph. Systems must predict "no answer" (often scored as the [CLS] token) when appropriate. SQuAD 2.0 remains a primary testbed for extractive QA with answerability.
- Natural Questions (2019): A large-scale Google dataset of real Google queries and Wikipedia pages. Annotators mark long and short answers or label a question NULL if no answer is found. Kwiatkowski et al. [39] report that about 50.5% of sampled queries had no relevant passage at all, and another ~14.7% had only partial answers, which they label as unanswerable. (On the official splits, roughly one-third of training examples are answerable.) NQ's format includes yes/no classification for questions answerable only with a Boolean, adding another dimension to answerability.

• NewsQA (2017): QA over CNN news articles. Questions were written using only article summaries, so many questions end up unanswerable in the full article. Trischler et al. [40] collected ~100K QA pairs; later analyses [36] note that a significant fraction (~13–14% in train) are "nil" questions with no answer span. This makes NewsQA a valuable benchmark for nil-aware extraction models.

• BoolQ (2019): A crowdsourced yes/no reading-comprehension dataset. Each question has a paragraph and a binary (yes/no) answer. By design, every question is answerable with a yes-or-no answer. However, BoolQ is often viewed as requiring answer verification or inference (i.e., entailment) rather than span extraction. It demonstrates that answerability can be viewed as a classification problem (yes/no) given context, related to but distinct from "no-answer" detection.

• QuAC (2018) and CoQA (2019): Conversational QA datasets. In QuAC, a student asks free-form questions about a hidden Wikipedia paragraph, and the teacher answers with spans or "n/a" (no answer). Many QuAC questions are unanswerable or require dialog context. CoQA similarly includes some unanswerable turns (annotated as "" or "unknown"). These tasks stress models' ability to track context and declare "no answer" when the conversation's current question cannot be answered from the text.

• TyDi QA (2020): A multilingual info-seeking QA dataset across 11 languages. In its primary tasks, systems must select the passage containing the answer (or NULL if none exists) and then produce an answer span or a yes/no response. Thus, TyDi explicitly trains models to output NULL or no answer when needed. Its "gold passage" variant, however, follows SQuAD's convention by discarding questions that are unanswerable from the given passage.

Transformer-based models, including BERT and T5, currently dominate QA systems. These models generally incorporate answerability in one of three ways: (i) multi-class output: extending the output space to include a "no answer" label (for example, adding a special token or classification head); (ii) confidence estimation: using calibrated probabilities or external classifiers to decide when to abstain; or (iii) answer verification: using a second-pass model to check an answer's validity. For instance, Hu et al. [38] demonstrate that a two-stage reader-verifier architecture can significantly improve handling of SQuAD 2.0. A key insight from recent analysis is that many unanswerable questions have distinctive cues. Yatskar [41] finds that SQuAD 2.0's unanswerable questions often involve "entity salad," false premises, or missing information, making them relatively easy to identify. QA datasets are known to contain annotation artifacts; Gururangan et al. [42] and others have shown that even without a passage, simple heuristics applied to the question alone can predict answerability or answers in some crowdsourced data. These artifacts can inadvertently help answerability models cheat, but also risk poor generalization. Kamath et al. [37] warned that models trained on SoftMax confidence may overfit to such artifacts and perform poorly under distribution shift.

They advocate calibration or the use of external signals to make the abstention decision more robust. In multilingual settings, answerability becomes more complex. Aside from TyDi QA, cross-lingual datasets (e.g., XOR QA) sometimes phrase answerability as retrieving the correct snippet in any language or answering "no" if no snippet is sufficient. Generally, the core issue is similar: models must learn to predict NULL/absence. Notably, many translation-based QA benchmarks (MLQA, XQuAD) drop unanswerable items by construction, so answerability is less emphasized there. Finally, a recent trend is the use of large generative models, such as GPT-style models. These models often produce answers even when users are not confident, raising safety concerns. Some work explores techniques for "self-verification" or uncertainty-aware generation. For example, one can prompt an LLM to say "I cannot answer" when uncertain explicitly. Preliminary studies on abstention in LLMs indicate that explicit training or thresholding can enable models to withhold answers when confidence is low. This is an active area of research, especially for open-domain QA.

### 2.3 *Generation of Synthetic Data for QA and Text Relations Modeling*

Deep generative models are increasingly used to produce synthetic text data for NLP tasks. These methods can greatly expand limited training corpora by "generate-annotate-learn" approaches [43]. For instance, synthetic question-answer (QA) pairs and paraphrase sentences can be generated by fine-tuned or prompted language models, then labeled or filtered to train classifiers. Synthetic datasets serve both as additional training data and as challenging evaluation sets that probe model behavior, such as adversarial or domain-generalization benchmarks. For example, Takahashi et al. [44] utilize an instruction-tuned model to generate Japanese QA pairs, achieving performance comparable to that of human-curated data. Likewise, Hosseini et al. [45] create a high-quality synthetic NLI (entailment/contradiction) dataset spanning multiple domains, which significantly improves cross-domain accuracy. Overall, synthetic data aims to augment scarce labels, enable domain transfer, and improve robustness in low-resource settings [44,46].

#### 2.3.1 Synthetic QA Datasets

Synthetic data has been used extensively to augment QA systems. Early work [47] introduced "round-trip consistency" QA generation, showing improvements on SQuAD and Natural Questions. More recent systems train on millions of automatically generated QA pairs [48]. Shakeri et al. [49] propose an end-to-end BART model that, given a passage, generates (answer, question) pairs; they train on SQuAD and then fine-tune on target domains, significantly improving cross-domain QA. Hematian Hemati and Beigy [50] address conversational QA by incorporating synthetic follow-up questions into the dialogue history, thereby making models more robust to augmented conversation turns.

In multilingual and low-resource settings, synthetic QA is especially valuable. Takahashi et al. [44] synthesize Japanese QA pairs using an English-trained instructive large language model (LLM) and find that the model fine-tuned on this synthetic data matches human-curated performance. Namboori [46] develops GeMQuAD, utilizing an AlexaTM 20B LLM with one-shot prompts to generate QA pairs for Hindi and Spanish, and then applies a semi-supervised Weak-DAP filtering approach to select high-quality examples. These examples show that prompt-based generation can overcome the scarcity of labeled data.

Selecting and filtering synthetic QA is crucial. Jin and Wang [52] note that naively adding all generated QA can harm performance if the quality is poor. They propose using a large LLM as a "reward model" in reinforcement learning to pick the best synthetic QA pairs. Other filtering methods include using the generation probability [47] or an external QA model to evaluate the quality of generated questions. In all cases, synthetic QA for training is typically combined with real data to avoid model drift. For evaluation, synthetic QA has also been used to stress-test systems (e.g., adversarial questions), though most studies focus on training augmentation.

#### 2.3.2 Synthetic Paraphrase Datasets

Paraphrase detection and generation also benefit from synthetic examples. One approach is neural question paraphrasing, where Moon and Fan [53] generate paraphrase pairs of questions by combining answer-aware question generation (QG) with filtering. They demonstrate that their QG-based paraphrases outperform simple methods, such as back-translation or synonym replacement. In low-resource languages, synthetic paraphrases can be created via "pivot" or MLM techniques.

Akil et al. [49] construct BanglaParaphrase by first generating a raw synthetic set (via bilingual pivoting and MLM) and then filtering with lexical and semantic metrics. They mask selected Bangla tokens, feed them into BanglaBERT to generate variants, and retain only pairs that exceed the PINC and BERTScore thresholds to ensure diversity and meaning. The result is a high-quality paraphrase corpus that improves model training.

Other augmentation methods include simple data-augmentation heuristics. For instance, Wei and Zou [54] proposed EDA (Easy Data Augmentation), which uses synonym replacement and random swaps to improve classification performance. Such strategies can generate paraphrase-like text cheaply, though they lack the linguistic fidelity of neural methods. More recently, large pretrained models have been prompted to rephrase sentences directly. Across all approaches, the goal is to enrich models' exposure to alternate phrasings without additional human labeling. Metrics such as BERTScore [55] and PINC [56] are often used to filter out trivial or low-quality paraphrases in synthetic datasets.

#### 2.3.3 Synthetic NLI and Entailment Datasets

Synthetic generation for Natural Language Inference (NLI) and textual entailment is a newer area. Hosseini et al. [45] find that creating a diverse NLI dataset (GNLI) through large language model (LLM) generation can significantly improve domain generalization. They prompt models to generate premises and hypotheses across various domains, ensuring that the synthetic examples have accurate entailment labels. This yields NLI pairs formed in creative ways, rather than through trivial edits, with high label precision.

Other work uses synthetic data to probe or evaluate NLI. For example, the HANS dataset [57] is a synthetic evaluation set designed to test models on lexical heuristics. Although not generated by neural models, it illustrates the value of synthetic test cases. Generative approaches to NLI, such as chain-of-thought prompts or multi-step pipelines, can yield premise–hypothesis–label triples [45]. In related tasks, Tang et al. [58] generate synthetic claim-passage pairs for fact-checking using multi-stage LLM prompts, an approach analogous to natural language inference (NLI) generation. Overall, synthetic NLI aims to cover unseen domains and linguistic phenomena (e.g., negation, quantifiers) that extend beyond existing datasets, such as SNLI or MultiNLI.

#### 2.3.4 Quality Control Strategies

Ensuring the quality of synthetic text is critical. Common strategies include:

• *Filtering by Model Confidence*: Use generation probabilities as scores. For example, Shakeri et al. [59] use the likelihood of the generated QA (from BART) to filter out low-confidence pairs. Alberti et al. [48] similarly require that synthetic QA be answerable by a pretrained model in a round-trip check.

• *Metric-based Filtering*: Compute lexical/semantic metrics between the original and generated text. In paraphrase datasets, thresholds on PINC and BERTScore [55] ensure paraphrases are both diverse and meaning-preserving. For QA generation, answer overlaps and language quality measures (such as BLEU and ROUGE) may be used.

• *Classifier or LLM Evaluation*: He et al. [43] advocate using a strong classifier to label synthetic examples: the synthetic text is first generated (or labeled) and then passed through the "best available" task model to obtain pseudo-labels. Jin and Wang [51] train an RL-based selector with an LLM reward to pick high-quality QA pairs, outperforming naive selection. In other words, models themselves act as gatekeepers on generated data.

• *Human Verification*: When feasible, humans can vet a sample of synthetic data to calibrate filters. Akil et al. [49] use human evaluation to set BERTScore thresholds for Bangla paraphrases. This ensures the chosen thresholds align with actual semantic correctness.

• *Mix with Real Data*: Crucially, studies note that synthetic examples should supplement rather than replace human data. He et al. [43] and others caution that iterative augmentation should retain original data to avoid feedback loops. As long as the gold data remains in training, synthetic augmentation can steadily improve models without divergence.

By combining these quality control tactics, researchers mitigate noise in synthetic datasets. In practice, a multi-stage pipeline is often employed, involving the generation of candidate examples, the application of automated filters (such as likelihood, metrics, and model checks), and the optional use of a small set of human ratings to adjust thresholds. These steps help ensure that synthetic QA, paraphrase, and NLI examples are coherent, relevant, and beneficial for learning.

## 3. Materials and Methods

### *3.1 Synthetic STR Dataset Construction*

As a first step, we evaluate each passage with our baseline model (GPT-4.1) and retain only those for which the model correctly answers at least **k** questions from the original SQuAD dataset (Table 1). For passages with more than **k** answerable questions, we normalize them to contain precisely k validated questions, each correctly answerable by GPT-4.1, by trimming the set. Empirically, setting **k** = 6 provides sufficient informational diversity to construct meaningful STR distinctions while still allowing controlled, interpretable information manipulation in subsequent steps.

In the second step, we generate a paraphrased version of each passage using the GPT-4.1 model, prompting it to alter the wording and surface form while preserving the underlying information content. This process yields a pair of semantically equivalent texts despite having different lexical and syntactic expressions. We again use GPT-4.1 to verify that the paraphrased text preserves the same answerable question set as the original. Any paraphrase that does not maintain this equivalence is discarded and excluded from further processing. We use the METEOR metric to ensure that the generated paraphrases are not trivial restatements of the original text. Specifically, we apply a lower METEOR threshold ($t<0.6$) to filter out paraphrases that remain too similar in surface form, retaining only those that achieve sufficient lexical and syntactic divergence while preserving the underlying information content.

We generate variants of both the original and paraphrased texts by selectively removing questions from their AQS in a controlled manner. In practice, each text is submitted to the baseline model, along with precise instructions to remove the information corresponding to a chosen subset of questions (n = 0, 1, 2, 3, 4, 5), while retaining the information required to answer the remaining k-n questions. This produces aligned variants whose answerable question sets differ systematically and in an interpretable way.

We perform an additional validation and post-filtering round to ensure that each generated text strictly matches its AQS. All answerable questions must be answerable from the text, and all questions marked as unanswerable must remain unanswerable. This round is executed using a dedicated GPT-4.1 prompt. *Appendix A* provides a few examples to illustrate the dataset construction process.

Using the resulting text variants and their associated AQSs, we construct STR pairs and assign labels based on the set relations between their AQSs. We control the balance between relation types by specifying the number of samples drawn for each category. Specifically, for each paraphrase source, we generate $m_{\text{equivalence}} = 6$, $m_{\text{inclusion}} = 10$, and $m_{\text{overlap}} = 20$ labeled pairs. This produces a consistent and balanced distribution across the three STR categories for proper model training and evaluation.

**Table 1.** An example SQuAD 2.0 SQuAD 1.1 context with associated questions and the corresponding synthetic texts in the STR dataset

| *Original(A)*: The Apollo program was the third United States human spaceflight program carried out by NASA, which accomplished the first human landing on the Moon from 1969 to 1972. First conceived during Dwight D. Eisenhower's administration as a follow-up to Project Mercury, which put the first Americans in space, Apollo was later dedicated to President John F. Kennedy's national goal of "landing a man on the Moon and returning him safely to the Earth" by the end of the 1960s. | | |
|---|---|---|
| **Question** | | **Answer** |
| Q1 | Which space agency was responsible for the Apollo program? | *NASA* |
| Q2 | What was the goal of the Apollo program? | *Landing a man on the Moon and returning him safely to the Earth* |
| Q3 | When did the Apollo Moon landings take place? | *from 1969 to 1972* |
| Q4 | Who initiated the Apollo program? | *Dwight D. Eisenhower* |
| Q5 | Which program came before Apollo? | *Project Mercury* |
| *Paraphrasing(B):* Emerging during the administration of Dwight D. Eisenhower, as the project intended to succeed Project Mercury—which had placed the first Americans in space—the Apollo program became the third U.S. human spaceflight initiative carried out by NASA. It was later aligned with President John F. Kennedy's objective of "landing a man on the Moon and returning him safely to the Earth" before the 1960s ended, ultimately achieving the first human lunar landing between 1969 and 1972 | | |
| *Rephrased with only Q5 unanswerable(C):* The Apollo program was NASA's third crewed spaceflight initiative and achieved humanity's first Moon landing, operating from 1969 to 1972. Initially formulated during the administration of President Dwight D. Eisenhower, it was later aligned with President John F. Kennedy's objective of "landing a man on the Moon and returning him safely to Earth" before the close of the 1960s. | | |
| *Rephrased with only Q1 unanswerable(D):* The Apollo program was the third U.S. crewed spaceflight effort and achieved the first human landing on the Moon between 1969 and 1972. It was initially formulated during President Dwight D. Eisenhower's tenure as a successor to Project Mercury, which sent the first Americans into orbit. The program was later aligned with President John F. Kennedy's objective of "landing a man on the Moon and returning him safely to the Earth" before the decade's end. | | |
| *Labels:* Equivalence*:* A=B; *Inclusion:* C≤A,C≤B,D≤A,D≤*B;* Overlap*:* A ⋈ B | | |

### 3.2 *STR classification model training*

Using the constructed dataset, we train and evaluate several STR classification models.

Generative language models can perform multi-class STR classification using prompt-based zero-shot and few-shot learning. The task is to assign each text pair to one of three semantic relation types: equivalence (both texts express the same information), inclusion (one text fully contains the other's information and adds more), and semantic overlap (the texts share some content but neither fully contains the other).

In this setup, classification is framed as a generative task: the model receives a natural-language instruction and must output the correct label for a given text pair. In the zero-shot setting, the prompt defines the task and lists the available labels. In the few-shot setting, the prompt includes several labeled examples per class, providing in-context supervision and enabling analogical generalization.

We evaluate two model variants, GPT-4.1 and GPT-4o, under both zero-shot and few-shot configurations, with and without chain-of-thought (CoT) prompting. In the CoT setup, the model first produces a short rationale before generating the final label, encouraging more explicit reasoning for this challenging semantic task. *Appendix B* provides prompts used for zero-shot and few-shot STR classification using generative models.

We also train and evaluate several transformer-based discriminative models for direct multi-class STR classification. We explore two approaches:

(1) fine-tuning transformer-based cross-encoders, where both texts are fed jointly into a single model that performs sentence-pair classification, and

(2) using frozen sentence encoders, where the embeddings of both texts obtained from a pre-trained SBERT model are concatenated and used as features for classical classifiers.

For the cross-encoder setting, we evaluate several transformer architectures, including Longformer-base, RoBERTa-base, and DistilBERT, all fine-tuned end-to-end for sentence-pair classification. The two input texts are concatenated using model-specific separator tokens (e.g., [CLS] text1 [SEP] text2 [SEP] in BERT), and the contextual embedding of the [CLS] token is passed through a classification head to predict one of the three STR labels. All models are fine-tuned with cross-entropy loss for three epochs, with a batch size of six.

In the frozen-encoder setting, we train Random Forest and Logistic Regression classifiers using features extracted from a fixed SBERT backbone. These classifiers operate on sentence embeddings and incorporate simple auxiliary text-based features (such as part-of-speech tag distributions or Jaccard similarity between texts) to improve prediction accuracy.

## 4. Results

We evaluated seven different approaches for classifying SCR between document pairs, reporting performance using Accuracy and macro-average F1 as primary metrics. Table 3 presents the comprehensive results of our classification experiments.

**Table 2.** Classification Results for Semantic Coverage Relation Detection.

| Model | Accuracy | Macro-F1 |
|---|---|---|
| RoBERTa-base | **0.614** | 0.446 |
| DistilBERT | 0.606 | 0.441 |
| SBERT + Logistic Regression | 0.604 | **0.478** |
| SBERT + Random Forest | 0.591 | **0.529** |
| Longformer-base | 0.555 | 0.238 |
| GPT-4.1 Zero-Shot | 0.339 | 0.341 |
| GPT-4.1 Few-Shot | 0.313 | 0.307 |
| GPT-4.1 Zero-Shot, CoT | 0.408 | 0.409 |
| GPT-4.1 Few-Shot CoT | 0.406 | 0.398 |
| GPT-4o Zero-Shot | 0.354 | 0.355 |
| GPT-4o Few-Shot | 0.311 | 0.265 |
| GPT-4o Zero-Shot CoT | 0.399 | 0.406 |
| GPT-4o Few-Shot CoT | 0.279 | 0.220 |

The results reveal several significant findings about the relative effectiveness of different modeling approaches for semantic coverage relation detection:

Fine-tuned cross-encoder models significantly outperformed generative approaches. RoBERTa-base achieved the highest overall accuracy at 61.4%, followed closely by DistilBERT at 60.6%. This demonstrates the effectiveness of task-specific fine-tuning for nuanced semantic classification tasks. Pretrained sentence-encoder with classical ML-based classification heads showed competitive performance with better class balance. Notably, Random Forest achieved the highest macro-F1 score of 52.9%, indicating superior performance across all three relation classes. Logistic Regression also performed well with a macro-F1 of 47.8%, demonstrating that carefully engineered features can effectively capture semantic coverage patterns.

Generative models struggled with the task. All GPT variants performed poorly, with accuracy ranging from 27.9% to 40.8%. The baseline GPT-4.1 zero-shot achieved 33.9% accuracy, while GPT-4.1 few-shot achieved 31.3% accuracy. These results suggest that this fine-grained semantic classification task requires specialized training rather than general language understanding capabilities.

CoT prompting showed mixed results. While CoT prompting improved some GPT-4.1 variants (zero-shot improved from 33.9% to 40.8%), it had inconsistent effects across different models and settings, and GPT-4o few-shot with justification performed worst at 27.9% accuracy.

The confusion matrices (*Appendix C*) further reveal distinct patterns in how different models classify the three relation types.

*The equivalence relation proved most challenging for all models, which misclassified most* equivalent pairs as semantic overlap. This is evident from the confusion matrices, which show that most equivalent-relation examples are misclassified as semantic overlap, likely because subtle semantic preservation is required when texts exhibit substantial lexical variation.

*Inclusion relations* were handled most effectively by Random Forest-based and other discriminative models, contributing to their superior macro-F1 scores. The hierarchical nature of decision trees appears well-suited to capturing the asymmetric nature of the inclusion relation.

*Semantic overlap* relation showed the most consistent performance across different model types, with most models correctly identifying the majority of semantic overlap cases. This indicates that partial overlapping patterns are somewhat easier to detect, possibly because this class is the most common in the dataset.

The RoBERTA-base model demonstrated the most robust overall performance, benefiting from its optimized pre-training regimes and larger training corpora compared to standard BERT. Its success validates the importance of high-quality pre-trained representations for semantic understanding tasks.

## 5. Conclusion, Limitations, and Future Work

In this work, we introduced the explainable Semantic Text Relation (STR) framework, a fine-grained approach for analyzing semantic relations between pairs of texts. We proposed a three-class typology: Equivalence, Inclusion, and Semantic Overlap.

Drawing on question-answering frameworks used for factual consistency and other tasks related to semantic text analysis, we represent a text's meaning through its Answerable Question Set (AQS). We then define STR as a set-theoretic relation between the AQS of two texts.

Because each relation is determined by which questions are answerable or unanswerable from each text, the framework provides inherent explainability: the predicted label is directly traceable to explicit question-answerability evidence. To support the development and evaluation of STR models, we constructed a systematic synthetic dataset derived from the SQuAD 2.0 corpus using GPT-4.1 for controlled paraphrasing and information omission.

As a first step, we trained and evaluated several models for STR type classification (without explanation). These included generative models with few-shot and CoT prompting, fine-tuned transformer-based cross-encoders, and classical machine-learning models using frozen pre-trained transformer embeddings.

The constructed dataset provides not only STR labels for training and evaluating classification models but also ground-truth AQS annotations for assessing explanation models. Future work can naturally focus on developing and evaluating explanation methods and joint classification–explanation models.

The proposed methodology for creating a synthetic STR dataset is fully generalizable. Because it relies on paraphrasing, controlled information removal, and AQS-based semantic labeling, the same process can be applied to a wide range of QA datasets. Any dataset that provides question–answer pairs can be transformed into controlled text variants with predictable answerability patterns, enabling automatic construction of STR labels. This opens the door to generating richer and more diverse benchmarks across domains, languages, and task settings, supporting robust training and evaluation of STR classification and explanation models.

In this work, we provided an initial evaluation of several model families using basic performance metrics on the STR dataset. Next steps might include a more thorough assessment of the models with additional metrics. Moreover, more advanced methods for training STR models can be explored to improve inference accuracy, including experimenting with a broader range of model architectures and optimized training hyperparameters.

Future work may also investigate incorporating the STR dataset into standard benchmarking suites to enable more systematic evaluation of general-purpose LLMs on STR-related tasks.

## Appendix A: Datapoint Example

*A.1 Source Paragraph*

*The New York City Fire Department (FDNY) provides fire protection, technical rescue, response to various hazards, and emergency medical services throughout the five boroughs of New York City. It is the largest municipal fire department in the United States and the second largest in the world, after the Tokyo Fire Department. The FDNY employs approximately 11,080 uniformed firefighters and over 3,300 uniformed EMTs and paramedics. The department is often referred to as New York's Bravest.*

*A.2 Questions (AQS)*

1. *What are the primary responsibilities and operational areas of the FDNY?*
2. *How does the FDNY's size compare to other fire departments in the U.S. and globally?*

3. *What is the approximate number of uniformed personnel in the FDNY?*
4. *What is the well-known motto or nickname associated with the department?*

*A.3 Full Paraphrase*

*The Fire Department of New York (FDNY) is responsible for fire suppression, technical rescue operations, initial response to biological, chemical, and radiological incidents, and emergency medical care throughout all five boroughs of New York City. As the largest municipal fire department in the United States, the FDNY ranks second globally, behind only the Tokyo Fire Department. The department has a workforce of around 11,080 uniformed firefighters and more than 3,300 uniformed emergency medical technicians and paramedics. The department is commonly known by its motto, "New York's Bravest".*

*A.4 Synthetically generated text with answer to Q4 removed*

Removed question*: What is the well-known motto or nickname associated with the department?*

Generated text:

*The Fire Department of New York (FDNY) is responsible for fire suppression, technical rescue operations, initial response to biological, chemical, and radiological incidents, and emergency medical care throughout all five boroughs of New York City. As the largest municipal fire department in the United States, the FDNY ranks second globally, behind only the Tokyo Fire Department. The department has a workforce of around 11,080 uniformed firefighters and more than 3,300 uniformed emergency medical technicians and paramedics.*

## Appendix B: Prompts for STR Classification using Generative Models

*B.1 Zero-shot Prompt*

*You are a language expert tasked with identifying the semantic relation between two texts. The possible relations are: EQUIVALENCE – Both texts express the same information. INCLUSION – one text contains all the information in the other, plus additional content. SEMANTIC OVERLAP – the texts have partial semantic overlap, but neither fully includes the other.*
*Text A: "{TEXT A}" Text B: "{TEXT B}"*
*What is the semantic relation between Text A and Text B?*
*Answer with one of: "EQUIVALENCE", "INCLUSION", or "SEMANTIC OVERLAP".*

*B.2 Few-shot Prompt*

*You are a language expert tasked with identifying the semantic relation between two texts. The possible relations are: EQUIVALENCE – Both texts express the same information. INCLUSION – one text contains all the information in the other, plus additional content. SEMANTIC OVERLAP – the texts have partial semantic overlap, but neither fully includes the other.*
*Example 1: Text A: "The Eiffel Tower is located in Paris and attracts millions of tourists every year." Text B: "Many tourists visit the Eiffel Tower in Paris annually." Answer: INCLUSION*
*Example 2:*
*Text A: "Photosynthesis occurs in plant leaves using sunlight, water, and carbon dioxide." Text B: "The process of photosynthesis in plants uses water, CO, and sunlight in leaves." Answer: EQUIVALENCE*
*Example 3: Text A: "The collapse of mortgage-backed securities triggered the 2008 financial crisis." Text B: "The Great Depression was caused by a stock market crash in 1929." Answer: SEMANTIC OVERLAP*
*Now, determine the relation in the following example: Text A:*
*"{TEXT A}" Text B: "{TEXT B}"*
*Answer*

## Appendix C: Confusion Matrices for STR classification models

**Figure 1.** Confusion matrices for GPT-4.1 and GPT-4o in zero-shot and few-shot settings, with and without chain-of-thought (CoT) justification prompting, for Semantic Text Relation (SCR) classification.

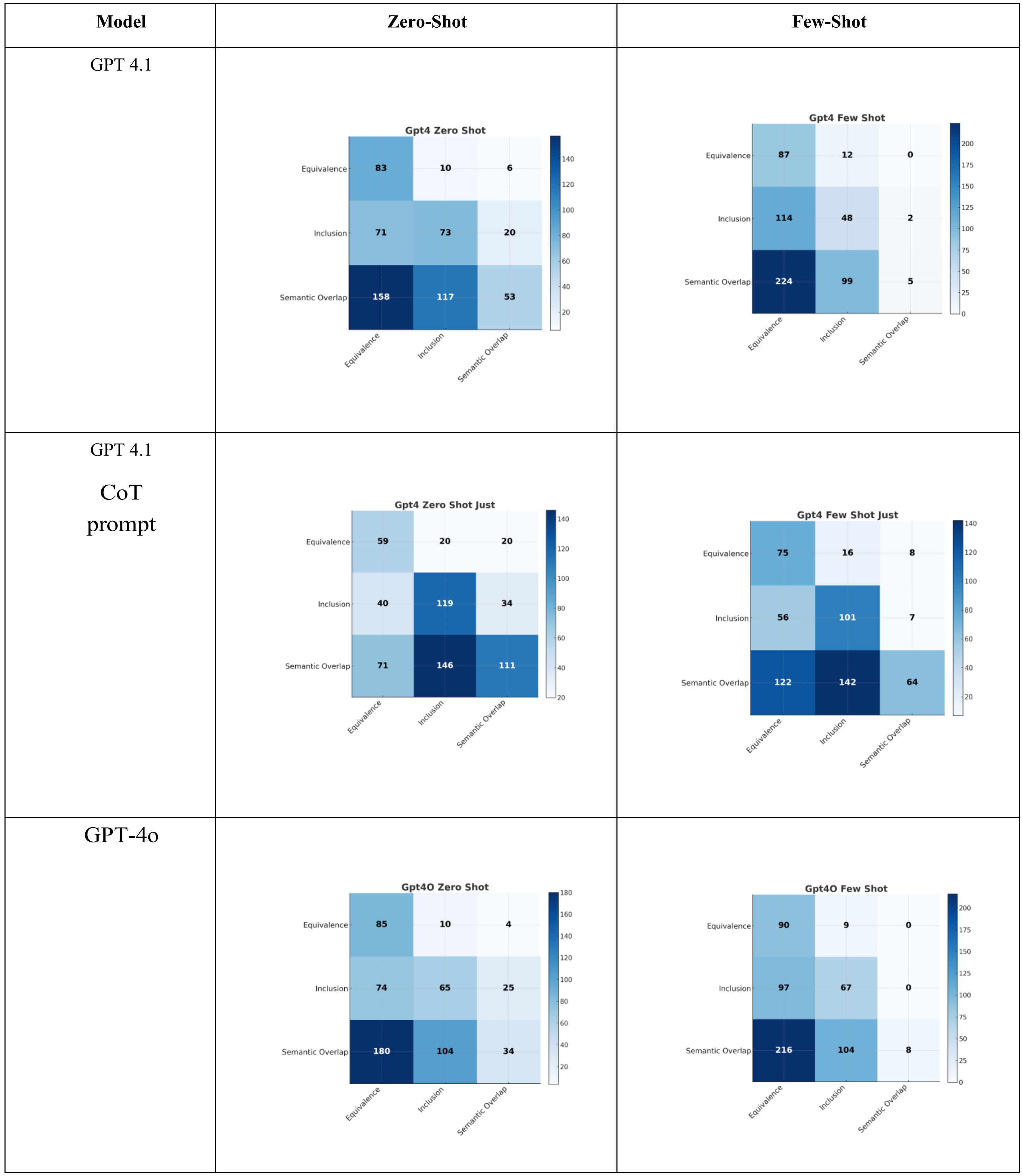

| Model | Zero-Shot | Few-Shot |
| --- | --- | --- |
| GPT 4.1 | Gpt4 Zero Shot | Gpt4 Few Shot |
| GPT 4.1 CoT prompt | Gpt4 Zero Shot Just | Gpt4 Few Shot Just |
| GPT-4o | Gpt4O Zero Shot | Gpt4O Few Shot |

GPT-4o

CoT prompt

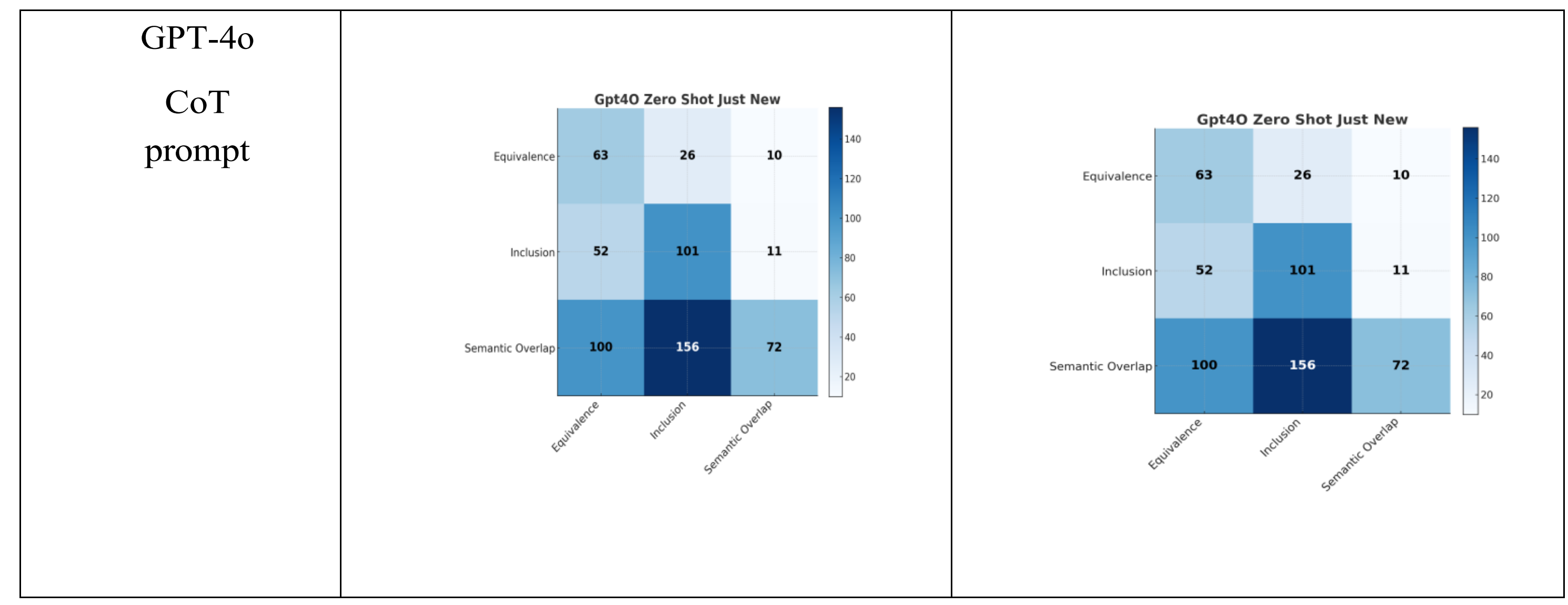